\documentclass[10pt,twocolumn,letterpaper]{article}

\usepackage{cvpr}
\usepackage{times}
\usepackage{epsfig}
\usepackage{graphicx}
\usepackage{amsmath}
\usepackage{amssymb}
\usepackage{booktabs}
\usepackage{algorithm}
\usepackage{algorithmicx}
\usepackage{varwidth}
\usepackage{multirow}
\usepackage{algpseudocode}

\def\ie{{\em i.e.}}

\usepackage{verbatim}
\usepackage[breaklinks=true,bookmarks=false]{hyperref}

\cvprfinalcopy 

\def\cvprPaperID{****} 

\ifcvprfinal\fi
\begin{document}


\title{FeatherNets: Convolutional Neural Networks as Light as Feather \\for Face Anti-spoofing}
\author{Peng Zhang$^{\rm 2,3}$, Fuhao Zou$^{\rm 1}$\thanks{Corresponding author}, Zhiwen Wu$^{\rm 2}$, Nengli Dai$^{\rm 3}$\\
Skarpness Mark$^{\rm 2}$, Michael Fu$^{\rm 2}$, Juan Zhao$^{\rm 2}$, Kai Li$^{\rm 1}$\\
\tt\footnotesize $^{\rm 1}$School of Computer Science and Technology, Huazhong University of Science and Technology\\
\tt\footnotesize$^{\rm 2}$Intel IAGS SSP\\
\tt\footnotesize$^{\rm 3}$Wuhan National Laboratory for Optoelectronics, Huazhong University of Science and Technology\\
\tt\footnotesize
\{peng3zhang,fuhao\_zou,dainl\}@hust.edu.cn\\
\tt\footnotesize
\{peng3.zhang,zhiwen.wu,juan.j.zhao,michael.fu,mark.skarpness\}@intel.com
}
\maketitle
\thispagestyle{empty}

\begin{abstract}
Face Anti-spoofing gains increased attentions recently in both academic and industrial fields. With the emergence of various CNN based solutions, the multi-modal(RGB, depth and IR) methods based CNN showed better performance than single modal classifiers. However, there is a need for improving the performance and reducing the complexity. Therefore, an extreme light network architecture(FeatherNet A/B) is proposed with a streaming module which fixes the weakness of Global Average Pooling and uses less parameters. Our single FeatherNet trained by depth image only, provides a higher baseline with \textbf{0.00168 ACER}, \textbf{0.35M parameters and 83M FLOPS}. Furthermore, a novel fusion procedure with ``ensemble + cascade'' structure is presented to satisfy the performance preferred use cases. Meanwhile, the MMFD dataset is collected to provide more attacks and diversity to gain better generalization. We use the fusion method in the Face Anti-spoofing Attack Detection Challenge@CVPR2019 and got the result of \textbf{0.0013(ACER), 0.999(TPR@FPR=10e-2), 0.998(TPR@FPR=10e-3) and 0.9814(TPR@FPR=10e-4)}.
\end{abstract}
\section{Introduction}

Currently, face recognition is an important way for identity authentication systems. However, it confronts with the challenge caused by face spoofing attacks such as the 2D/3D Presentation Attack. Therefore, it is important to equip the system with robust anti-spoofing algorithms. Anti-spoofing is usually regarded as a problem of binary classification. Some works are texture-based using binary classifiers with handcrafted features\cite{maatta2011face,komulainen2013context,de2013can,boulkenafet2017face}. However, these methods suffer from poor generalization because the texture information varies with cameras/capture devices. Another problem of texture-based approaches is that the texture information is not as discriminative as the depth information on task of 2D presentation attack detection.

The depth information is more discriminative since the depth of the real face is uneven, and the depth images of the attacking face is plane. Atoum \textit{et al.} \cite{atoum2017face} exploited the depth supervised procedure. Nevertheless, the depth information
is estimated from RGB image and not as accurate as the depth image captured by depth camera such as RealSense 300\footnote{ https://realsense.intel.com/}.

Recently, deep learning techniques are widely used to extract deep features\cite{feng2016integration,li2016original,patel2016cross}, which have richer semantical information compared to traditional handcrafted features. Hence utilizing the deep learning for face PAD has been widely used recently. 

However, there is a new trend that face recognition is gradually moving to the mobile devices or embedded devices. This requires the face anti-spoofing algorithms to run with 
less computation and storage costs. From this perspective, the design of deep learning based anti-spoofing algorithms become more challenge in the mobile or embedded environments. Thus, it is necessary to develop a light-weight deep learning algorithm so that spoofing detection can be used.

\begin{figure*}[t]
	\begin{center}
 \includegraphics[scale=0.35]{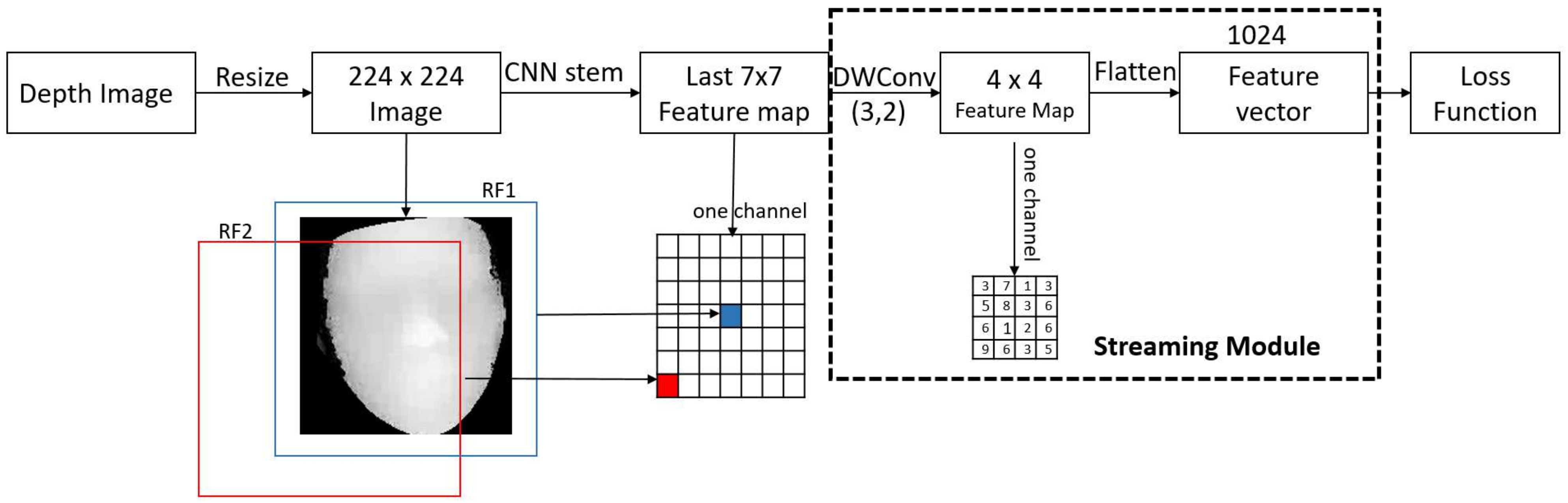}
	\end{center}
	\caption{Our depth faces feature embedding CNN structure. In the last 7$\times$7 feature map, the receptive field and the edge (RF2) portion of the middle part (RF1) is different, because their importance is different. DWConv is used instead of the GAP layer to better identify this different importance. At the same time, the fully connected layer is removed, which makes the network more portable.}
	\label{fig:delgap}
\end{figure*}
To address the issues of computational and storage costs, we design a light-weight CNN architecture (named as FeatherNet) which gets a higher accuracy and computational complexity. Firstly, FeatherNets have a thin CNN stem, thus the computational cost is less. Secondly, a new architecture (named as Streaming Module) is proposed, which has better performance in terms of accuracy than the Global Average Pooling (GAP) approach.

We also design a new fusion classifier architecture which assembles and cascades several models learned from multi-modal data, \ie{}, the depth and IR data, to generate better prediction accuracy than single depth models. Although the depth image is discriminative on 2D presentation attack detection, multi-modal fusion can boost the performance further due to its complementary and generalization capability\cite{zhang2018casia}.
The new fusion procedure has been applied to face anti-spoofing competition@CVPR2019 and showed the result of 0.0013 (ACER), 0.999 (TPR@FPR=10e-2), 0.998 (TPR@FPR=10e-3) and 0.9814 (TPR@FPR=10e-4) in the test dataset.

The major contributions of this paper are summarized as follows:
\textbf{a)}.An extremely light CNN architecture with a \textit{Streaming Module} which has good performance;
\textbf{b)}.A novel \textit{fusion procedure} with ``ensemble+cascade” structure which outperforms the single model classifiers;
\textbf{c)}.A new Multi-Modal Face
Dataset (\textit{MMFD}) is collected which will be released recently and a new data augmentation algorithm is applied on training.


\section{Related work}
The related work is reviewed in two categories in chronological order: traditional and CNN based methods.

\textbf{Traditional}: Face anti-spoofing is treated as a binary classification problem by traditional SVM (Support Vector Machine), through two steps as below:

1) Crafted features detection: Various filters were used to detect the points to present the feature. The widely adopted features include: 
 Local Binary Patterns (LBP) \cite{de2012lbp,de2013can,maatta2011face},
Scale Invariant Feature Transform (SIFT)\cite{patel2016secure}, 
 Speeded-Up Robust Features (SURF)\cite{boulkenafet2017face}, 
histogram of oriented gradients (HOG)\cite{komulainen2013context,yang2013face}, 
Difference of Gaussian (DoG)\cite{yang2013face}.

2) Liveness or not classification through SVM or Random Forest\cite{chakraborty2014overview}.

However, Wang \textit{et al.} \cite{wang2018exploiting} indicated that the feature detection is  greatly influenced by the environment, for example the lighting condition. Furthermore, the feature detection shows limited features, and the feature points don't provide as many features' information as those CNN methods could bring with the huge data sets. 

\textbf{CNN based}: There are mainly three types of CNN based PAD. 

1) Using RGB single frame with binary supervision\cite{li2016original,patel2016cross}:
 Most approaches just adopt the final fully-connected layer to distinguish the real and fake faces. While Li \textit{et al.} \cite{li2016original} proposed a way to link the deep partial features (from CNN) and Principle Component Analysis (PCA) to reduce the dimension, and lastly they used SVM to distinguish real and fake faces.
 Patel \textit{et al.} \cite{patel2016cross} applied the action features (such as eye blinking) to enhance the state of the art. And the researchers found that this can still be improved through multiple supervisions.
 
2)Using RGB multi-frame with depth or rPPG (remote photoplethysmography) supervision\cite{atoum2017face,hernandez2018time}: Two different types of supervision are applied: depth or rPPG. Different frames are also captured by the shift of camera and frames to anticipate the depth. Moreover, researchers analyzed the presentation attack and video-based pulse detection. Live Faces could show some blood signal through rPPG but not in fake face. Recently, Liu \textit{et al.} \cite{Liu2018CVPR} proposed a procedure which uses single frame to regression depth map and uses multi-frame to predict rPPG, which is a good way to distinguish living face. This network architecture combining CNN and RNN, could simultaneously estimate the depth map and rPPG signal of the face.

3)Recently, Zhang \textit{et al.} \cite{zhang2018casia} provided a large-scale multi-modal dataset, namely CASIA-SURF, which consists of 3 modalities data (RGB, depth and IR). It provides a strong baseline to make full use of these features by fusing multi-modal data through a three-stream network.

There are two main aspects to enhance for the multi-modal method: (1) The baseline performance of CASIA-SURF still has a lot of room to improve; (2) The adoption of light-weight network architecture that can benefit more edge side applications. In next section, an extreme lite network architecture is proposed which uses depth and IR information as supervision respectively to learn complementary models, achieving a well trade-off between performance and computational burden. Furthermore, a novel fusion classifier with ``ensemble + cascade'' structure is proposed for the performance preferred use case.

\section{Approach}

In this section, we will introduce the details of the FeatherNets. Inspired by the equal importance gap of Global Average Pooling (GAP) in face tasks, a new Streaming module is adopted in the FeatherNets which can provide a strong baseline for Face Anti-spoofing. Furthermore, to achieve higher performance, the ``ensemble + cascade'' fusion procedure will be proposed.

\subsection{FeatherNet Architecture Design}

The existing anti-spoofing networks\cite{patel2016cross,li2016original,hernandez2018time,wang2018exploiting} have the problems of large parameters and weak generalization ability. For this reason, FeatherNets architecture is proposed, targeting a network as lite as feather.

\subsubsection{The Weakness of GAP for Face Task}
~~Global Average Pooling (GAP) is employed by a lot of state-of-the-art networks for object recognition task, e.g.ResNets\cite{he2016deep}, DenseNet\cite{huang2017densely} and some light-weight networks, like  MobilenetV2\cite{sandler2018mobilenetv2}, Shufflenet\_v2\cite{ma2018shufflenet}, IGCV3\cite{sun2018igcv3}. GAP has been proved on its ability of reducing dimensions and preventing over-fitting for the overall structure\cite{lin2013network}. However, for the face related tasks, Wu\cite{wu2018shift} and Deng\cite{deng2018arcface} have observed that CNNs with GAP layer are less accurate than those without GAP. Meanwhile, MobileFaceNet\cite{chen2018mobilefacenets} replaces the GAP with Global Depthwise Convolution (GDConv) layer, and explains the reason why it is effective through the theory of receptive field\cite{long2014convnets}. The main point of GAP is "equal importance" which is not suitable for face tasks. 

As shown in Figure \ref{fig:delgap}, the last 7 $\times$ 7 feature map is denoted as FMap-end, each cell in FMap-end corresponds to a receptive field at different position. The center blue cell corresponds to RF1 and the edge red one corresponds to RF2. As described in\cite{luo2016understanding}, the distribution of impact in a receptive field distributes as a Gaussian, the center of a receptive field has more impact on the output than the edge. Therefore, RF1 has larger effective receptive field than RF2. For our face anti-spoofing task, the network input is 224 $\times$ 224 images which only contain the face region. As above analysis, the center unit of FMap-end is more important than the edge one. GAP is not applicable to this case. One choice is to use fully connected layer instead of GAP, this will introduce large number of parameters to the whole model and increase the risk of over-fitting.

\subsubsection{Streaming Module}

\begin{figure}[htb]
\begin{center}
\includegraphics[width=0.99\linewidth]{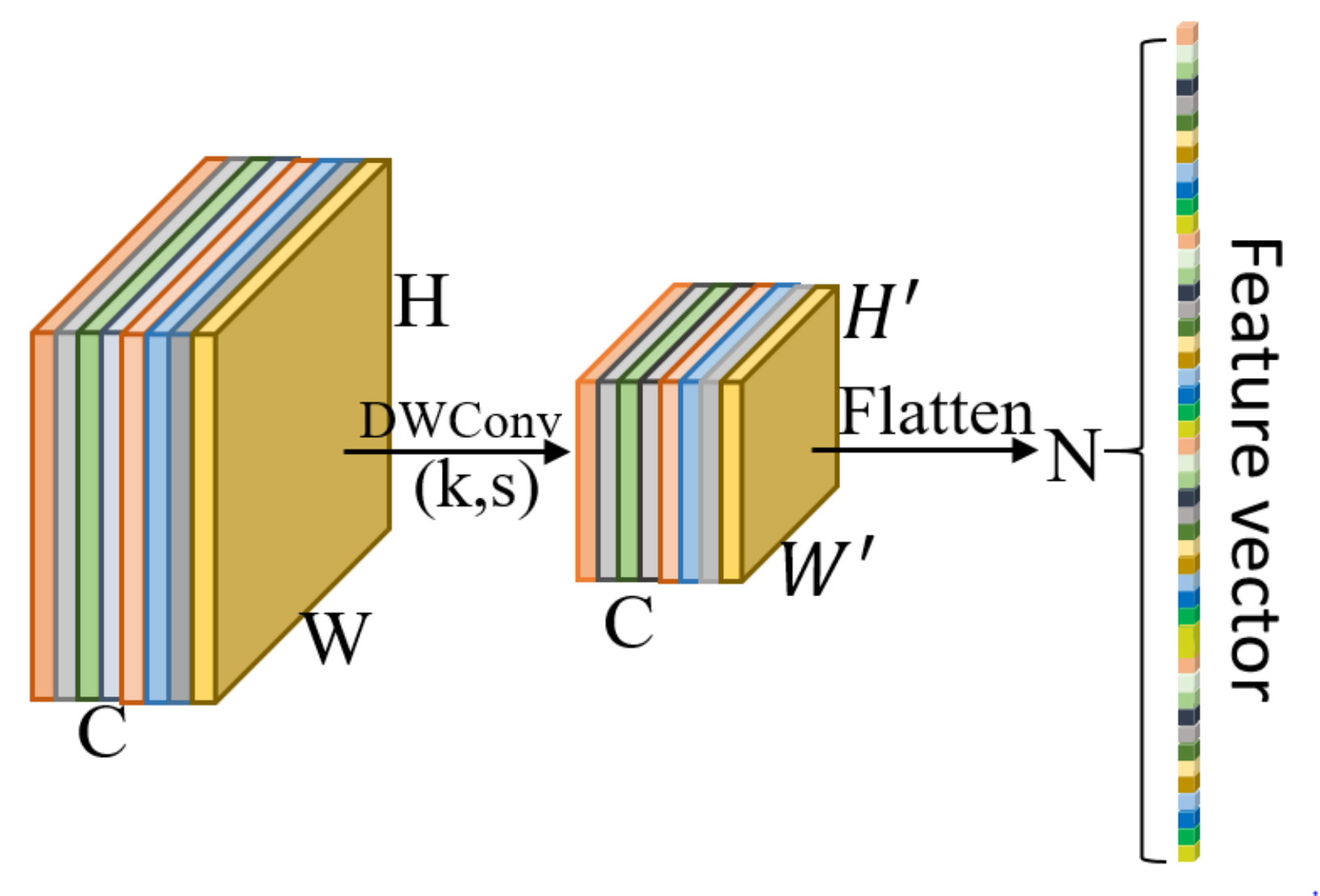}
\end{center}
   \caption{Streaming Module. The last blocks' output is down-sampled by a depthwise convolution\cite{chollet2017xception,howard2017mobilenets} with stride larger than 1 and flattened directly into an one-dimensional vector.}
\label{fig:DCF}
\end{figure}

\begin{figure*}[t]
	\centering
	\includegraphics[scale=0.4]{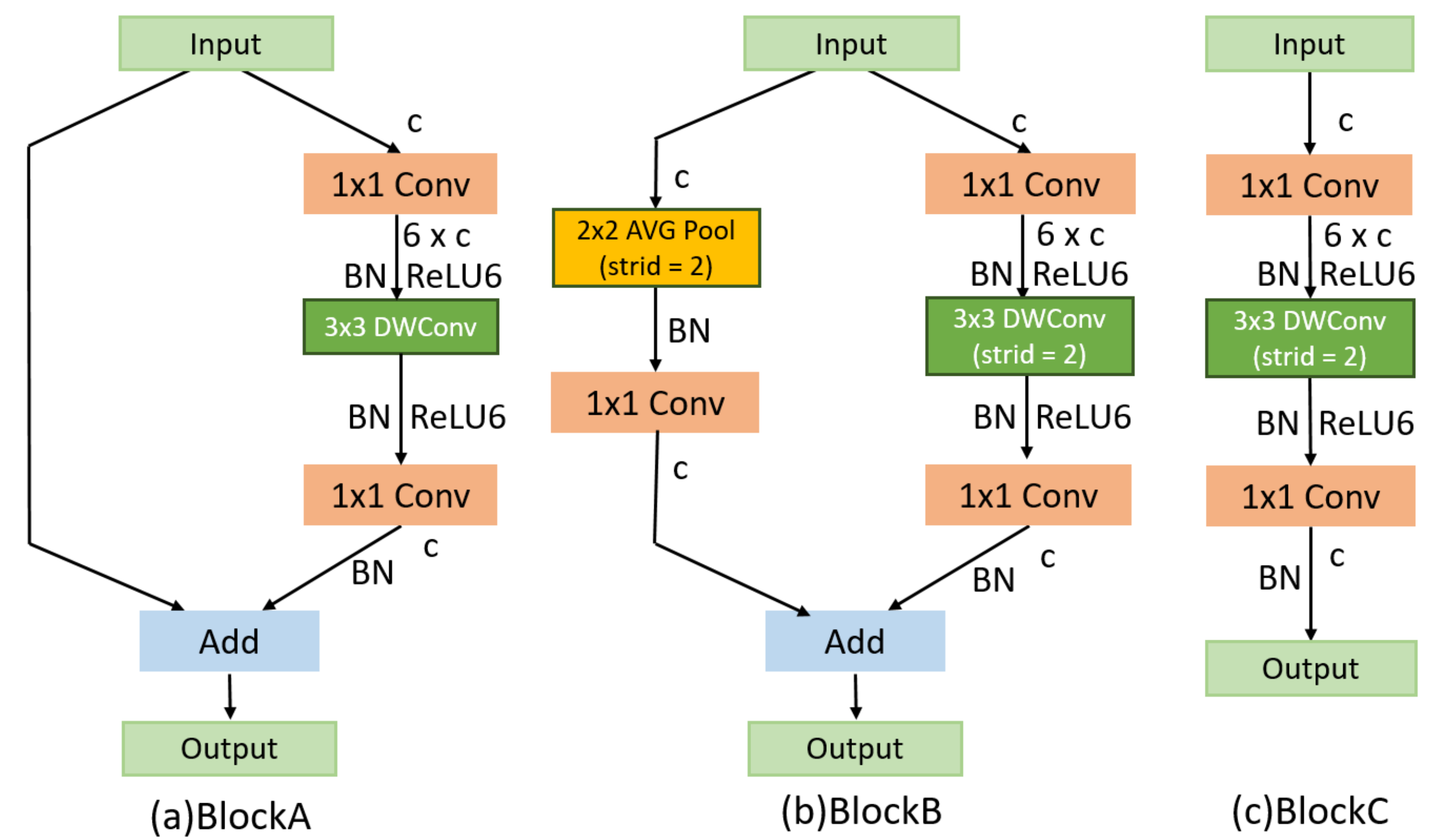}
	\caption{FeatherNets' main blocks. FeatherNetA includes BlockA \& BlockC. FeatherNetB includes BlockA \& BlockB. (BN: BatchNorm; DWConv: depth wise convolution; c:number of input channels.)}
	\label{fig:module}
\end{figure*}

~~To treat different units of FMap-end with different importance, Streaming Module is designed, as shown in the Figure \ref{fig:DCF}. In Streaming Module, a depthwise convolution (DWConv) layer with stride larger than 1 is used  for down-sampling whose output, is then flattened directly into an one-dimensional feature vector. The compute process is represented by equation (\ref{equ:Depthwise}).
\begin{equation}
FV_{n(y,x,m)}=\sum_{i,j} K_{i,j,m}\cdot F_{IN_{y}(i), IN_{x}(j), m}
\label{equ:Depthwise}
\end{equation}
In equation \ref{equ:Depthwise}, FV is the flattened feature vector while $N$ = $H^{'} \times W^{'} \times C$ elements ($H^{'}$, $W^{'}$ and $C$ denote the height, width and channel of DWConv layer's output feature maps respectively). $n(y, x, m)$, computed as equation (\ref{equ:nyxm}), denotes the $n_{th}$ element of FV which corresponds to the $(y, x)$ unit in the $m_{th}$ channel of the DWConv layer's output feature maps.
\begin{equation}
\label{equ:nyxm}
n(y,x,m) = m \times H^{'} \times W^{'} + y \times H^{'}+x
\end{equation}
On the right side of the equation (\ref{equ:Depthwise}), K is the depthwise convolution kernel, F is the FMap-end of size H$\times$W$\times$C (H, W and C denote the height, width and channel of FMap-end respectively). m denotes the channel index. i,j denote the spatial position in kernel K, and $IN_{y}(i)$, $IN_{x}(j)$ denote the corresponding position in F. They are computed as equation (\ref{equ:iny}), (\ref{equ:inx}).
\begin{equation}
\label{equ:iny}
IN_{y}(i) = y \times S_{0} + i
\end{equation}
\begin{equation}
\label{equ:inx}
IN_{x}(j) = x \times S_{1} + j
\end{equation}

$S_{0}$ is the vertical stride and $S_{1}$ is the horizontal stride.
A fully connected layer is not added after flattening feature map, because this will increase more parameters and the risk of overfitting. Meanwhile, related experiments are processed to verify the reason for removing the fully connected layer, as show in Table \ref{table:ablations}.

Streaming module can be used to replace global average pooling and fully connected layer in traditional networks.

\subsubsection{Network Architecture Detail}
~~Besides Streaming Module, there are BlockA/B/C as shown in Figure \ref{fig:module} to compose FeatherNetA/B. The detailed structure of the primary FeatherNet architecture is shown in Table \ref{table:architectures}. 
\textbf{BlockA} is the inverted residual blocks proposed in  MobilenetV2\cite{sandler2018mobilenetv2}. BlockA is used as our main building block which is shown in the Figure \ref{fig:module}(a). The expansion factors are the same as in  MobilenetV2\cite{sandler2018mobilenetv2} for blocks in our architecture. 
\textbf{BlockB} is the down-sampling module of FeatherNetB. Average pooling (AP) has been proved in Inception\cite{szegedy2015going} to benefit performance, because of its ability of embedding multi-scale information and aggregating features in different receptive fields. Therefore, average pooling (2 $\times$ 2 kernel with stride = 2) is introduced in BlockB (Figure \ref{fig:module}(b)). Besides, in the network ShuffleNet\cite{ma2018shufflenet}, the down-sampling module joins $3\times3$ average pooling layer with stride=2 to obtain excellent performance. Li \textit{et al.} \cite{xie2018bag} suggested that increasing average pooling layer works well and impacts the computational cost little. Based on the above analysis, adding pooling on the secondary branch can learn more diverse features and bring performance gains. The performance comparison between using the auxiliary branch (BlockB in Figure \ref{fig:module}(b)) and not using the branch (BlockC in Figure\ref{fig:module}(c)) is showing in the Table \ref{table:ablations}.
\textbf{BlockC} is the down-sampling Module of our network FeatherNetA. BlockC is faster and with less complexity than BlockB. According to our experiment in Table \ref{table:performance}, FeatherNetA used less parameters.
\begin{table}[!http]
\centering
\begin{tabular}{c|c|c|c}
\midrule
Input&Operator&t&c\\
\hline
224$^2$$\times$3&Conv2d,/2&-&32\\
112$^2$$\times$32&BlockB&1&16\\
56$^2$$\times$16&BlockB&6&32\\
28$^2$$\times$32&BlockA&6&32\\
28$^2$$\times$32&BlockB&6&48\\
14$^2$$\times$48&5xBlockA&6&48\\
14$^2$$\times$48&BlockB&6&64\\
7$^2$$\times$64&2xBlockA&6&64\\
\hline
7$^2$$\times$64&Streaming&-&1024\\
\bottomrule 
\end{tabular}
\caption{Network Architecture: FeatherNet B. All spatial convolutions use 3 $\times$ 3 kernels. The expansion factor t is always applied to the input size, while c means number of Channel. Meanwhile, every stage SE-module\cite{hu2018squeeze} is inserted with reduce = 8. And FeatherNetA replaces BlockB in the table with BlockC.}
\label{table:architectures}
\end{table}

\begin{figure*}
\centering
\includegraphics[scale=0.3]{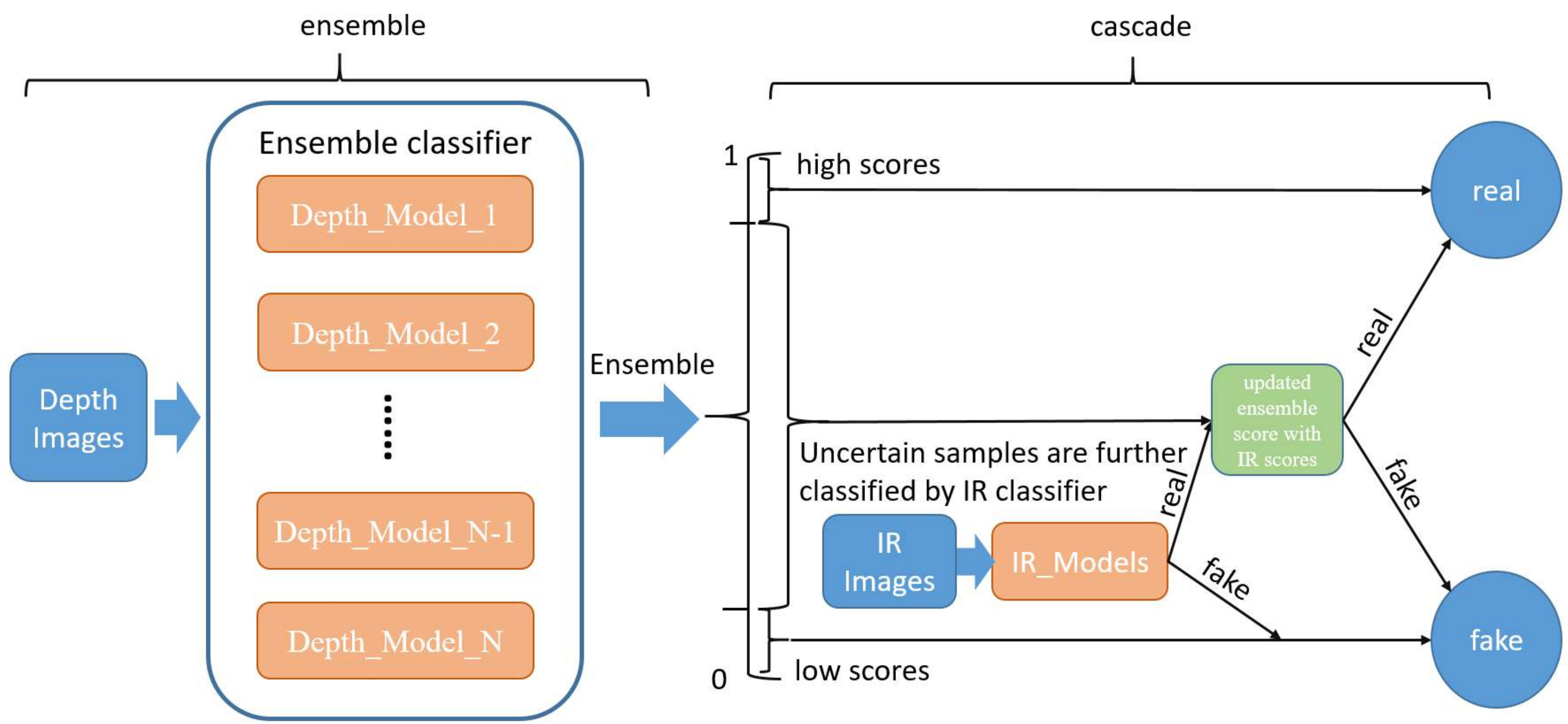} 
\caption{Multi-Modal Fusion Strategy: Two stages cascaded, stage 1 is an ensemble classifier consisting of several depth models. Stage 2 employs IR models to classify the uncertain samples from stage 1.}
\label{fig:ensemble_cascade}
\end{figure*}

After each down-sampling stage, SE-module\cite{hu2018squeeze} is inserted with reduce = 8 in both FeatherNetA and FeatherNetB. In addition, when designing the model, a fast down-sampling strategy\cite{qin2018fd} is used at the beginning of our network which makes the feature map size decrease rapidly and without much parameters.
Adopting this strategy can avoid the problem of weak feature embedding and high processing time caused by slow down-sampling due to limited computing budget\cite{duong2018mobiface}. The primary FeatherNet only has 0.35M parameters.

The FeatherNets' structure is built on BlockA/B/C as mentioned above except for the first layer which is a fully connected. As shown in Table \ref{table:architectures}, the size of the input image is 224 $\times$ 224. A layer with regular convolutions, instead of depthwise convolutions, is used at the beginning to keep more features. Reuse channel compression to reduce 16 while using inverted residuals and linear bottleneck with expansion ratio = 6 to minimize the loss of information due to down-sampling.
Finally, the Streaming module is used without adding a fully connected layer, directly flatten the $4 \times 4 \times 64$ feature map into an one-dimensional vector, reducing the risk of over-fitting caused by the fully connected layer. After flattening the feature map, focal loss is used directly for prediction. The related ablation experiments are shown in the Table \ref{table:ablations}. When we added the fully connected layer, the performance dropped.

\subsection{Multi-Modal Fusion Method}\label{sec:MMFM}
The main idea for the fusion method is to use cascade inference on different modals: depth images and IR images. The model trained based on depth data could provide a high baseline (approximately 0.003 ACER in test set). According to our experiments, the IR data could provide a good performance in fake judgement for those samples that depth modal is not sure about. The cascade structure has two stages, as show in the Figure \ref{fig:ensemble_cascade}:\\
\textbf{Stage 1}: An ensemble classifier, consisting of multiple models , is employed to generate the predictions. These models are trained on depth data and from several checkpoints of different networks,  including FeatherNets. If the weighted average of scores from these models is near 0 or 1, input sample will be classified as fake or real respectively. Otherwise, the uncertain samples will go through the second stage.\\
\textbf{Stage 2}: FeatherNetB learned from IR data will be used to classify the uncertain samples from stage 1. The fake judgement of IR model is respected as the final result. For the real judgement, the final scores are decided by both stage 1 and IR models.

\section{Experiments}
The preliminary work will be introduced firstly, such as the evaluation metrics, datasets used for training, the proposed data augmentation method, the training settings of the FeatherNets and the baseline models. Secondly, the performance of the trained models (including FeatherNets) will be showed. Thirdly, the comparative experiments are used to show the validity of the MMFD dataset. Finally, the effectiveness of the network design is verified by ablation experiments.
\subsection{Preliminary Work}
\subsubsection{Evaluation Metrics}
~~For the performance evaluation, the following commonly used metrics\cite{komulainen2013context} will be introduced: Attack Presentation Classification Error Rate (APCER), Normal Presentation Classification Error Rate (NPCER) and Average Classification Error Rate (ACER). ACER is treated as the evaluation metric, in which APCER and NPCER are used to measure the error rate of fake or real samples, respectively.  Besides, the other metrics\cite{zhang2018casia} are also used, such as TPR@FPR=10E-2, 10E-3, 10E-4.
\subsubsection{Datasets}
~~Two datasets are used in the experiments: CASIA-SURF\cite{zhang2018casia} and the proposed Multi-Modal Face Dataset (MMFD).

\begin{figure}
\centering
\includegraphics[scale=0.4]{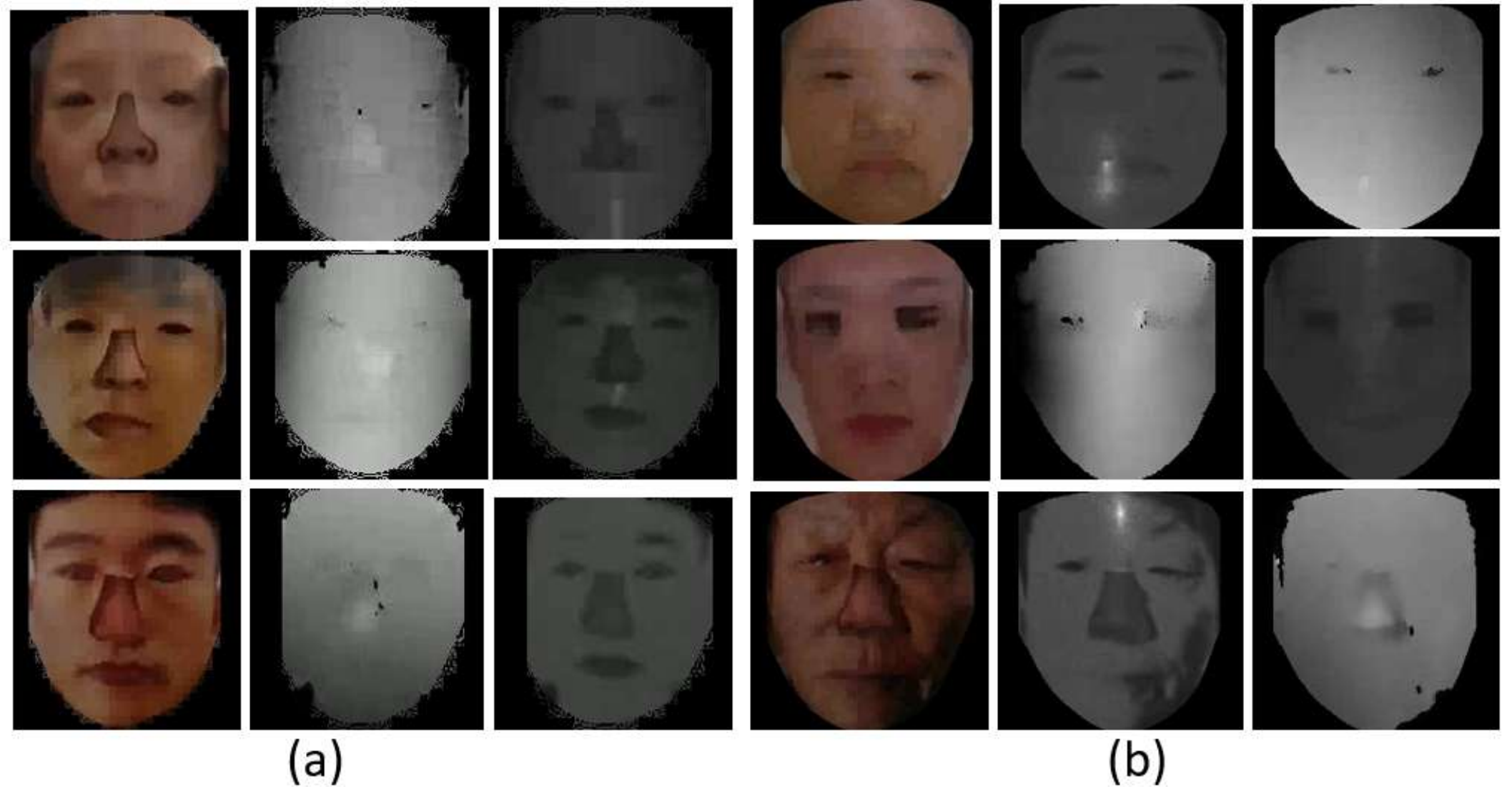}
\caption{(a)Training set contains attacks 4,5,6 (b)Validation and test sets contains attacks 1,2,3}.
\label{fig:datatype}
\end{figure}

\textbf{CASIA-SURF} is the largest publicly available dataset for face Anti-spoofing, provided by Surfing Technology\cite{zhang2018casia}. It consists of 1,000 subjects with 21,000 videos and each sample has 3 modalities (\ie, RGB, Depth and IR), as shown in Figure \ref{fig:datatype}. 
There are 6 attack ways of this dataset: 
\textit{Attack 1}: One person holds his/her flat face photo where eye regions are cut from the printed face.
\textit{Attack 2}: One person holds his/her curved face photo where eye regions are cut from the printed face.
\textit{Attack 3}: One person holds his/her flat face photo where eyes and nose regions are cut from the printed face.
\textit{Attack 4}: One person holds his/her curved face photo where eyes and nose regions are cut from the printed face.
\textit{Attack 5}: One person holds his/her flat face photo where eyes, nose and mouth regions are cut from the printed face.
\textit{Attack 6}: One person holds his/her curved face photo where eyes, nose and mouth regions are cut from the printed face.

\textbf{MMFD} In order to make the model more robust, more attack ways of diverse faces are collected. Then we sort out a dataset which is consisted of 15 subjects with 15415 real samples and 28438 fake samples, namely Multi-Modal Face Dataset (MMFD).

And each sample also has 3 modalities (RGB, Depth, IR). They are treated by the similar way as CASIA-SURF with a little modification. Besides the 6 attack ways of CASIA-SURF, 2 new attack ways are added. \textit{Attack A}: One person holds his/her flat face photo where eyes and mouth regions are cut from the printed face. \textit{Attack B}: One person holds his/her curved face photo where eyes and mouth regions are cut from the printed face. The presenters turn their head left/right/up/down to get different samples. Other variations on the presenters include: wearing glasses or not; opening mouth or not; moving face close to and far away from the camera; showing different emotions, e.g. happy, angry, sad and so on.

Collecting and masking steps are proposed to obtain the final images.
\textit{Collecting}: Intel RealSense SR300\footnote{ https://realsense.intel.com/} camera is used to generate RGB, Depth, IR and aligned-RGB frames simultaneously. RGB frame is $1280 \times 720$ resolution, Depth, IR and Aligned-Depth frames are $640 \times 480$ resolution.
\textit{Masking}: Dlib\cite{king2009dlib} is used to detect the bounding-box of face for RGB frame and Aligned-Depth frame. And the face region is passed into PRNet\cite{feng2018joint} to estimate the depth. To generate the mask image, the depth value of each pixel is checked in face box. If it is larger than 0.5, 1 will be sent otherwise 0 will be sent into mask image. At last, the RGB, Depth and IR images are multiplied with the mask, and only the face region is saved to files.
\begin{table*}[!ht]
	\begin{center}
		\centering
		\begin{tabular}{cccccc}
			\midrule
			Model&ACER&TPR@FPR=10E-2&TPR@FPR=10E-3&Params&FLOPS\\
			\hline
			ResNet18\cite{zhang2018casia}&0.05&0.883&0.272&11.18M&1800M\\
			Baseline\cite{zhang2018casia}&0.0213&0.9796&0.9469&--&--\\
			FishNet150(our impl)&0.00144&0.9996&0.998330&24.96M&6452.72M\\
			MobilenetV2(1)(our impl)&0.00228&0.9996&0.9993&2.23M&306.17M\\
			ShuffleNetV2(1)(our impl)&0.00451&1.0&0.98825&1.26M&148.05\\
			FeatherNetA&0.00261&1.0&0.961590&0.35M&79.99M\\
			FeatherNetB&\textbf{0.00168}&\textbf{1.0}&\textbf{0.997662}&\textbf{0.35M}&\textbf{83.05M}\\
			\bottomrule 
		\end{tabular}
		\caption{Performance in validation dataset. Baseline is a way of fusing three modalities data (IR, RGB, Depth) through a three-stream network. Only depth data is used for training in the other networks. FeatherNetA and FeatherNetB have achieved higher performance with less parameters. Finally, the models are assembled to reduce ACER to 0.0.}
		\label{table:performance}
	\end{center}
\end{table*}
\subsection{Implementation Detail}
\subsubsection{Data Augmentation}
\begin{figure}[ht]
\centering
\includegraphics[scale=0.4]{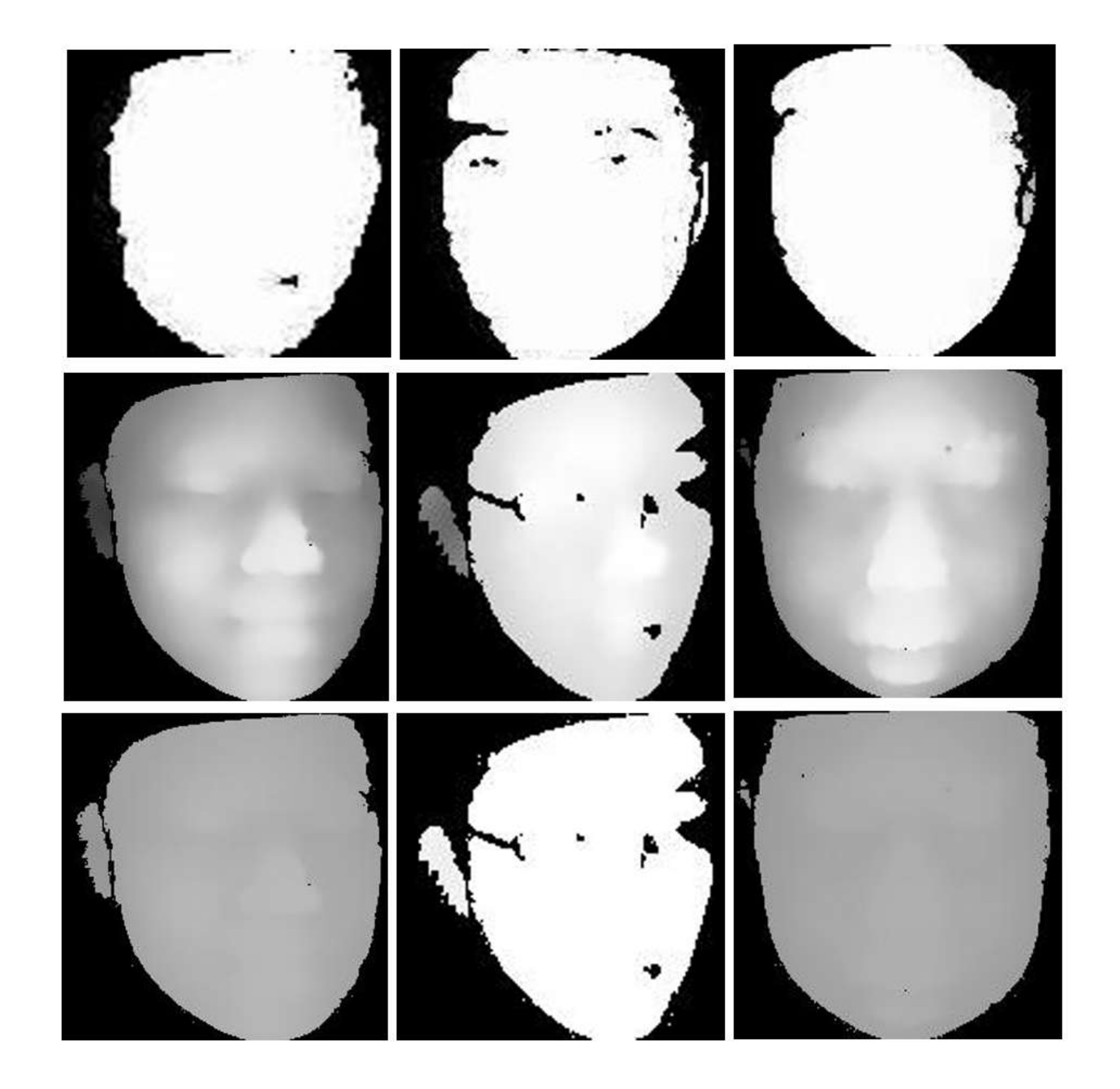}
\caption{depth image augmentation.(line 1): CAISA-SURF real depth images; (line 2): MMFD real depth images; (line 3): our augmentation method on MMFD.}
\label{fig:dataaugment}
\end{figure}
~~~~There are some differences in the images acquired by different devices, even if the same device model is used. As shown in the Figure \ref{fig:dataaugment}. The upper line is the depth images of the CASIA-SURF data set. The depth difference of the face part is small. It is difficult for the eyes to distinguish whether the face has a contour depth. The second line is the depth images of the MMFD dataset whose outline of the faces are clearly showed. In order to reduce the data difference caused by the device, the depth of the real face images is scaled in MMFD which can be seen in the third line of Figure \ref{fig:dataaugment}. The way of data augmentation is as Algorithm \ref{algo_depthaug}:

\begin{algorithm}[h]
\caption{Data Augmentation Algorithm}
\begin{algorithmic}[1]
 \State $scaler \gets$ a random value in range [1/8, 1/5]
 \State $offset \gets$ a random value in range [100, 200]
 \State $OutImg \gets 0$
 \For {$y = 0 \to Height-1$}
 \For {$x = 0 \to Width-1$}
 \If {$InImg(y,x) > 20$}
 \State $off \gets offset$
 \Else
 \State $off \gets 0$
 \EndIf
 \State $OutImg(y,x) \gets$ InImg(y,x) * scaler + off
 \EndFor
 \EndFor
 \State \Return{$OutImg$}
\end{algorithmic}
\label{algo_depthaug}
\end{algorithm}

\subsubsection{Training Strategy}
~~Pytorch\cite{paszke2017automatic} is used to implement the proposed networks. It initializes all convolutions and fully-connected layers with normal weight distribution\cite{he2015delving}. For optimization solver, Stochastic Gradient Descent(SGD) is adopted with both learning rate beginning at 0.001, and decaying 0.1 after every 60 epochs, and momentum setting to 0.9. The Focal Loss\cite{lin2017focal} is employed with $\alpha =1$ and $\gamma=3$.

\subsection{Result Analysis}
\subsubsection{How useful is MMFD dataset?}
~~~A comparative experiment is executed to show the validity and generalization ability of our data. 
As shown in Table \ref{table:generalization}, the ACER of FeatherNetB with MMFD depth data is better than that with CASIA-SURF\cite{zhang2018casia}, though only 15 subjects are collected. Meanwhile, the experiment shows that the best option is to train the network with both data. The results of using our FeatherNetB are much better than the baselines that use multi-modal data fusion, indicating that our network has better adaptability than the third-stream ResNet18 for baseline. 

\newcommand{\tabincell}[2]{\begin{tabular}{@{}#1@{}}#2\end{tabular}}
\begin{table}[!h]
\begin{center}
\centering
\begin{tabular}{ccc}
\midrule
Network&\tabincell{c}{Training Dataset}& ACER in Val\\
\hline
Baseline &CASIA-SURF&0.0213\\
FeatherNetB&CASIA-SURF depth& 0.00971\\
FeatherNetB&MMFD depth& 0.00677\\
FeatherNetB&\tabincell{c}{CASIA-SURF+\\MMFD depth}&\textbf{0.00168}\\
\bottomrule 
\end{tabular}
\caption{Performance of FeatherNetB training by different datasets. Column 3 means the ACER value in the validation dataset of CASIA-SURF\cite{zhang2018casia}. It shows that our dataset MMFD generalization ability is stronger than baseline of CASIA-SURF. The performance is better than the baseline method using multi-modal fusion.}
\label{table:generalization}
\end{center}
\end{table}

\subsubsection{Compare with other network performance}
~~~~As show in Table \ref{table:performance}, experiments are executed to compare with other network's performance. All experimental results are based on depth of CASIA-SURF and MMFD depth images, and then the performance is verified on the CASIA-SURF verification set. It can be seen from the table \ref{table:performance} that our parameter size is much smaller, only 0.35M, while the performance on the verification set is the best.

\subsubsection{Ablation Experiments}
~~~A number of ablations are executed to analyze different models with different layer combination, shown in Table \ref{table:ablations}. The models are trained with CASIA-SURF training set and MMFD dataset. 

\textbf{Why AP-down in BlockB}: Comparing \textit{Model1 and Model2}, Adding the Average Pooling branch to the secondary branch (called AP-down), as shown in block B of Figure \ref{fig:module}(b), can effectively improve performance with a small number of parameters.

\textbf{Why not use FC layer}: Comparing \textit{Model1 and Model3}, fully connected (FC) layer doesn't reduce the error when adding a fully connected layer to the last layer of the network. Meanwhile, a FC layer is computationally expensive. 

\textbf{Why not use GAP layer} Comparing \textit{Model3 and Model4}, it shows that adding global average pooling layer at the end of the network is not suitable for face anti-spoofing task. They will reduce performance. For more details, please refer to Section 3.

\begin{table}[!h]
\begin{center}
\centering
\begin{tabular}{ccccc}
\midrule
Model&FC&GAP&AP-down&ACER\\
\hline
Model1& $\times$ & $\times$ & $\times$ & 0.00261\\
Model2& $\times$ & $\times$\ & ${\checkmark}$ & \textbf{0.00168}\\
Model3& ${\checkmark}$ & $\times$\ & $\times$\ & 0.00325\\
Model4& ${\checkmark}$ & ${\checkmark}$ & $\times$\ & 0.00525\\
\bottomrule 
\end{tabular}
\caption{Ablation Experiments.}
\label{table:ablations}
\end{center}
\end{table}

\section{Competition details}
Based on CASIA-SURF\cite{zhang2018casia}, the Face Anti-spoofing challenge@CVPR2019 has been organized, aiming at compiling the latest efforts and research advances from the computational intelligence community in creating fast and accurate face spoofing detection algorithms\footnote{ http://chalearnlap.cvc.uab.es/workshop/32/description/}. This dataset provides a multi-modal dataset (RGB, Depth, IR) which is captured by Intel RealSense SR300. And it contains data for training, verification and the final evaluation. 

Our fusion procedure (described in section \ref{sec:MMFM}) is applied in this competition. Meanwhile, the proposed FeatherNets with depth data only can provide a higher baseline alone (around 0.003 ACER). During the fusion procedure, the selected models are with different statistic features, and can help each other. For example, one model's characteristics of low False Negative (FN) are utilized to further eliminate the fake samples. The detailed procedure is described as below:
\begin{algorithm}[ht]
	\caption{Ensemble Algorithm}
	\begin{algorithmic}[1]
		\State \begin{varwidth}[t]{\linewidth}
			$scores[] \gets$ 
			\par\hskip\algorithmicindent score\_FishNet150\_1, \par\hskip\algorithmicindent score\_FishNet150\_2, \par\hskip\algorithmicindent score\_ MobilenetV2,
			\par\hskip\algorithmicindent score\_FeatherNetA,
			\par\hskip\algorithmicindent score\_FeatherNetB, \par\hskip\algorithmicindent score\_ResNet\_GC
		\end{varwidth}
		\State $mean\_score \gets$ mean of scores[]
		\If {$mean\_score > max\_threshold\ ||\ mean\_score < min\_threshold$}
		\State $final\_score \gets mean\_score$
		\ElsIf {$score\_FishNet150\_1 < fish\_threshold$}
		\State $final\_score \gets score\_FishNet150\_1$
		\ElsIf {$score\_FeatherNetBForIR < IR\_threshold$}
		\State $final\_score \gets score\_FeatherNetBForIR$
		\Else
		\State \begin{varwidth}[t]{\linewidth} 
			$mean\_score \gets$
			\par\hskip\algorithmicindent (6 * mean\_score\ +\ score\_FishNet150\_1) / 7
		\end{varwidth}
		\If {$mean\_score > 0.$5}
		\State $final\_score \gets$ max of scores[]
		\Else
		\State $final\_score \gets$ min of scores[]
		\EndIf
		\EndIf
	\end{algorithmic}
	\label{algo_ensemble}
\end{algorithm}


\textbf{Training}: The depth data is used to train 7 models: FishNet150\_1, FishNet150\_2,  MobilenetV2, FeatherNetA, FeatherNetB, FeatherNetBForIR, ResNet\_GC. Meanwhile, FishNet150\_1, FishNet150\_2 are models from different epoch of FishNet. The IR data is used to train FeatherNetB as FeatherNetBforIR.

\textbf{Inference}: The inference scores will go through the ``ensemble + cascade'' process. The algorithm is shown as Algorithm \ref{algo_ensemble}.

\textbf{Competition Result}: The above procedure is used to get the result of 0.0013 (ACER), 0.999 (TPR@FPR=10e-2), 0.998 (TPR@FPR=10e-3) and 0.9814 (TPR@FPR=10e-4) in the test set and showed excellent performance in the Face Anti-spoofing challenge@CVPR2019.
\section{Conclusion}
We propose an extreme lite network architecture (FeatherNet A/B) with Streaming module, to achieve a well trade-off between performance and computational complexity for multi-modal face anti-spoofing. Furthermore, a novel fusion classifier with ``ensemble + cascade'' structure is proposed for the performance preferred use cases. Meanwhile, MMFD dataset is collected to provide more diverse samples and more attacks to gain better generalization ability. All these are used to join the Face Anti-spoofing Attack Detection Challenge@CVPR2019. The experiment and the competition results show that the proposed method can achieve excellent performance.
\section*{Acknowledgement}
The authors would like to thank Intel IAGS SSP DSS Video and Audio team\footnote{https://01.org/linuxmedia} members' great support and Web Platform Engineering team's help on Hardware devices support.

This work is supported in part by the National Natural Science Foundation of China under Grant No.61672254, 61672246, 61572221 and 61300222, Key project of National Natural Science Foundation of China Grant No U1536203, Natural Science Foundation of Hubei Province Grant No.2015CFB687, the Fundamental Research Funds for the Central Universities, HUST:2016YXMS088 and 2016YXMS018.
\clearpage

{\small
\bibliographystyle{unsrt}
\bibliography{egbib}

\begin{thebibliography}{10}

\bibitem{maatta2011face}
Jukka M{\"a}{\"a}tt{\"a}, Abdenour Hadid, and Matti Pietik{\"a}inen.
\newblock Face spoofing detection from single images using micro-texture
  analysis.
\newblock In {\em 2011 international joint conference on Biometrics (IJCB)},
  pages 1--7. IEEE, 2011.

\bibitem{komulainen2013context}
Jukka Komulainen, Abdenour Hadid, and Matti Pietik{\"a}inen.
\newblock Context based face anti-spoofing.
\newblock In {\em 2013 IEEE Sixth International Conference on Biometrics:
  Theory, Applications and Systems (BTAS)}, pages 1--8. IEEE, 2013.

\bibitem{de2013can}
Tiago de~Freitas~Pereira, Andr{\'e} Anjos, Jos{\'e}~Mario De~Martino, and
  S{\'e}bastien Marcel.
\newblock Can face anti-spoofing countermeasures work in a real world scenario?
\newblock In {\em 2013 international conference on biometrics (ICB)}, pages
  1--8. IEEE, 2013.

\bibitem{boulkenafet2017face}
Zinelabidine Boulkenafet, Jukka Komulainen, and Abdenour Hadid.
\newblock Face antispoofing using speeded-up robust features and fisher vector
  encoding.
\newblock {\em IEEE Signal Processing Letters}, 24(2):141--145, 2017.

\bibitem{atoum2017face}
Yousef Atoum, Yaojie Liu, Amin Jourabloo, and Xiaoming Liu.
\newblock Face anti-spoofing using patch and depth-based cnns.
\newblock In {\em 2017 IEEE International Joint Conference on Biometrics
  (IJCB)}, pages 319--328. IEEE, 2017.

\bibitem{feng2016integration}
Litong Feng, Lai-Man Po, Yuming Li, Xuyuan Xu, Fang Yuan, Terence Chun-Ho
  Cheung, and Kwok-Wai Cheung.
\newblock Integration of image quality and motion cues for face anti-spoofing:
  A neural network approach.
\newblock {\em Journal of Visual Communication and Image Representation},
  38:451--460, 2016.

\bibitem{li2016original}
Lei Li, Xiaoyi Feng, Zinelabidine Boulkenafet, Zhaoqiang Xia, Mingming Li, and
  Abdenour Hadid.
\newblock An original face anti-spoofing approach using partial convolutional
  neural network.
\newblock In {\em 2016 Sixth International Conference on Image Processing
  Theory, Tools and Applications (IPTA)}, pages 1--6. IEEE, 2016.

\bibitem{patel2016cross}
Keyurkumar Patel, Hu~Han, and Anil~K Jain.
\newblock Cross-database face antispoofing with robust feature representation.
\newblock In {\em Chinese Conference on Biometric Recognition}, pages 611--619.
  Springer, 2016.

\bibitem{zhang2018casia}
Shifeng Zhang, Xiaobo Wang, Ajian Liu, Chenxu Zhao, Jun Wan, Sergio Escalera,
  Hailin Shi, Zezheng Wang, and Stan~Z Li.
\newblock Casia-surf: A dataset and benchmark for large-scale multi-modal face
  anti-spoofing.
\newblock {\em arXiv preprint arXiv:1812.00408}, 2018.

\bibitem{de2012lbp}
Tiago de~Freitas~Pereira, Andr{\'e} Anjos, Jos{\'e}~Mario De~Martino, and
  S{\'e}bastien Marcel.
\newblock Lbp- top based countermeasure against face spoofing attacks.
\newblock In {\em Asian Conference on Computer Vision}, pages 121--132.
  Springer, 2012.

\bibitem{patel2016secure}
Keyurkumar Patel, Hu~Han, and Anil~K Jain.
\newblock Secure face unlock: Spoof detection on smartphones.
\newblock {\em IEEE Transactions on Information Forensics and Security},
  11(10):2268--2283, 2016.

\bibitem{yang2013face}
Jianwei Yang, Zhen Lei, Shengcai Liao, and Stan~Z Li.
\newblock Face liveness detection with component dependent descriptor.
\newblock In {\em 2013 International Conference on Biometrics (ICB)}, pages
  1--6. IEEE, 2013.

\bibitem{chakraborty2014overview}
Saptarshi Chakraborty and Dhrubajyoti Das.
\newblock An overview of face liveness detection.
\newblock {\em arXiv preprint arXiv:1405.2227}, 2014.

\bibitem{wang2018exploiting}
Zezheng Wang, Chenxu Zhao, Yunxiao Qin, Qiusheng Zhou, and Zhen Lei.
\newblock Exploiting temporal and depth information for multi-frame face
  anti-spoofing.
\newblock {\em arXiv preprint arXiv:1811.05118}, 2018.

\bibitem{hernandez2018time}
Javier Hernandez-Ortega, Julian Fierrez, Aythami Morales, and Pedro Tome.
\newblock Time analysis of pulse-based face anti-spoofing in visible and nir.
\newblock In {\em Proceedings of the IEEE Conference on Computer Vision and
  Pattern Recognition Workshops}, pages 544--552, 2018.

\bibitem{Liu2018CVPR}
Learning deep models for face anti-spoofing: Binary or auxiliary supervision.
\newblock In {\em In Proceedings of the IEEE Conference on Computer Vision and
  Pattern Recognition}, pages 389--398, 2018.

\bibitem{he2016deep}
Kaiming He, Xiangyu Zhang, Shaoqing Ren, and Jian Sun.
\newblock Deep residual learning for image recognition.
\newblock In {\em Proceedings of the IEEE conference on computer vision and
  pattern recognition}, pages 770--778, 2016.

\bibitem{huang2017densely}
Gao Huang, Zhuang Liu, Laurens Van Der~Maaten, and Kilian~Q Weinberger.
\newblock Densely connected convolutional networks.
\newblock In {\em Proceedings of the IEEE conference on computer vision and
  pattern recognition}, pages 4700--4708, 2017.

\bibitem{sandler2018mobilenetv2}
Mark Sandler, Andrew Howard, Menglong Zhu, Andrey Zhmoginov, and Liang-Chieh
  Chen.
\newblock Mobilenetv2: Inverted residuals and linear bottlenecks.
\newblock In {\em Proceedings of the IEEE Conference on Computer Vision and
  Pattern Recognition}, pages 4510--4520, 2018.

\bibitem{ma2018shufflenet}
Ningning Ma, Xiangyu Zhang, Hai-Tao Zheng, and Jian Sun.
\newblock Shufflenet v2: Practical guidelines for efficient cnn architecture
  design.
\newblock In {\em Proceedings of the European Conference on Computer Vision
  (ECCV)}, pages 116--131, 2018.

\bibitem{sun2018igcv3}
Ke~Sun, Mingjie Li, Dong Liu, and Jingdong Wang.
\newblock Igcv3: Interleaved low-rank group convolutions for efficient deep
  neural networks.
\newblock {\em arXiv preprint arXiv:1806.00178}, 2018.

\bibitem{lin2013network}
Min Lin, Qiang Chen, and Shuicheng Yan.
\newblock Network in network.
\newblock {\em arXiv preprint arXiv:1312.4400}, 2013.

\bibitem{wu2018shift}
Bichen Wu, Alvin Wan, Xiangyu Yue, Peter Jin, Sicheng Zhao, Noah Golmant, Amir
  Gholaminejad, Joseph Gonzalez, and Kurt Keutzer.
\newblock Shift: A zero flop, zero parameter alternative to spatial
  convolutions.
\newblock In {\em Proceedings of the IEEE Conference on Computer Vision and
  Pattern Recognition}, pages 9127--9135, 2018.

\bibitem{deng2018arcface}
Jiankang Deng, Jia Guo, Niannan Xue, and Stefanos Zafeiriou.
\newblock Arcface: Additive angular margin loss for deep face recognition.
\newblock {\em arXiv preprint arXiv:1801.07698}, 2018.

\bibitem{chen2018mobilefacenets}
Sheng Chen, Yang Liu, Xiang Gao, and Zhen Han.
\newblock Mobilefacenets: Efficient cnns for accurate real-time face
  verification on mobile devices.
\newblock In {\em Chinese Conference on Biometric Recognition}, pages 428--438.
  Springer, 2018.

\bibitem{long2014convnets}
Jonathan~L Long, Ning Zhang, and Trevor Darrell.
\newblock Do convnets learn correspondence?
\newblock In {\em Advances in Neural Information Processing Systems}, pages
  1601--1609, 2014.

\bibitem{luo2016understanding}
Wenjie Luo, Yujia Li, Raquel Urtasun, and Richard Zemel.
\newblock Understanding the effective receptive field in deep convolutional
  neural networks.
\newblock In {\em Advances in neural information processing systems}, pages
  4898--4906, 2016.

\bibitem{chollet2017xception}
Fran{\c{c}}ois Chollet.
\newblock Xception: Deep learning with depthwise separable convolutions.
\newblock In {\em Proceedings of the IEEE conference on computer vision and
  pattern recognition}, pages 1251--1258, 2017.

\bibitem{howard2017mobilenets}
Andrew~G Howard, Menglong Zhu, Bo~Chen, Dmitry Kalenichenko, Weijun Wang,
  Tobias Weyand, Marco Andreetto, and Hartwig Adam.
\newblock Mobilenets: Efficient convolutional neural networks for mobile vision
  applications.
\newblock {\em arXiv preprint arXiv:1704.04861}, 2017.

\bibitem{szegedy2015going}
Christian Szegedy, Wei Liu, Yangqing Jia, Pierre Sermanet, Scott Reed, Dragomir
  Anguelov, Dumitru Erhan, Vincent Vanhoucke, and Andrew Rabinovich.
\newblock Going deeper with convolutions.
\newblock In {\em Proceedings of the IEEE conference on computer vision and
  pattern recognition}, pages 1--9, 2015.

\bibitem{xie2018bag}
Junyuan Xie, Tong He, Zhi Zhang, Hang Zhang, Zhongyue Zhang, and Mu~Li.
\newblock Bag of tricks for image classification with convolutional neural
  networks.
\newblock {\em arXiv preprint arXiv:1812.01187}, 2018.

\bibitem{hu2018squeeze}
Jie Hu, Li~Shen, and Gang Sun.
\newblock Squeeze-and-excitation networks.
\newblock In {\em Proceedings of the IEEE conference on computer vision and
  pattern recognition}, pages 7132--7141, 2018.

\bibitem{qin2018fd}
Zheng Qin, Zhaoning Zhang, Xiaotao Chen, Changjian Wang, and Yuxing Peng.
\newblock Fd-mobilenet: Improved mobilenet with a fast downsampling strategy.
\newblock In {\em 2018 25th IEEE International Conference on Image Processing
  (ICIP)}, pages 1363--1367. IEEE, 2018.

\bibitem{duong2018mobiface}
Chi~Nhan Duong, Kha~Gia Quach, Ngan Le, Nghia Nguyen, and Khoa Luu.
\newblock Mobiface: A lightweight deep learning face recognition on mobile
  devices.
\newblock {\em arXiv preprint arXiv:1811.11080}, 2018.

\bibitem{king2009dlib}
Davis~E King.
\newblock Dlib-ml: A machine learning toolkit.
\newblock {\em Journal of Machine Learning Research}, 10(Jul):1755--1758, 2009.

\bibitem{feng2018joint}
Yao Feng, Fan Wu, Xiaohu Shao, Yanfeng Wang, and Xi~Zhou.
\newblock Joint 3d face reconstruction and dense alignment with position map
  regression network.
\newblock In {\em Proceedings of the European Conference on Computer Vision
  (ECCV)}, pages 534--551, 2018.

\bibitem{paszke2017automatic}
Adam Paszke, Sam Gross, Soumith Chintala, Gregory Chanan, Edward Yang, Zachary
  DeVito, Zeming Lin, Alban Desmaison, Luca Antiga, and Adam Lerer.
\newblock Automatic differentiation in pytorch.
\newblock 2017.

\bibitem{he2015delving}
Kaiming He, Xiangyu Zhang, Shaoqing Ren, and Jian Sun.
\newblock Delving deep into rectifiers: Surpassing human-level performance on
  imagenet classification.
\newblock In {\em Proceedings of the IEEE international conference on computer
  vision}, pages 1026--1034, 2015.

\bibitem{lin2017focal}
Tsung-Yi Lin, Priya Goyal, Ross Girshick, Kaiming He, and Piotr Doll{\'a}r.
\newblock Focal loss for dense object detection.
\newblock In {\em Proceedings of the IEEE international conference on computer
  vision}, pages 2980--2988, 2017.

\end{thebibliography}
}

\end{document}